\documentclass[lettersize,journal]{IEEEtran}
\usepackage{amsmath,amsfonts}
\usepackage{algorithmic}
\usepackage{algorithm}
\usepackage{array}
\usepackage[caption=false,font=normalsize,labelfont=sf,textfont=sf]{subfig}
\usepackage{textcomp}
\usepackage{stfloats}
\usepackage{url}
\usepackage{verbatim}
\usepackage{graphicx}
\usepackage{cite}

\usepackage{tikz}
\usepackage{pgfplots} 

\usepackage{scalerel}
\usetikzlibrary{svg.path}
\definecolor{orcidlogocol}{HTML}{A6CE39}
\tikzset{
    orcidlogo/.pic={
        \fill[orcidlogocol] svg{M256,128c0,70.7-57.3,128-128,128C57.3,256,0,198.7,0,128C0,57.3,57.3,0,128,0C198.7,0,256,57.3,256,128z};
        \fill[white] svg{M86.3,186.2H70.9V79.1h15.4v48.4V186.2z}
        svg{M108.9,79.1h41.6c39.6,0,57,28.3,57,53.6c0,27.5-21.5,53.6-56.8,53.6h-41.8V79.1z M124.3,172.4h24.5c34.9,0,42.9-26.5,42.9-39.7c0-21.5-13.7-39.7-43.7-39.7h-23.7V172.4z}
        svg{M88.7,56.8c0,5.5-4.5,10.1-10.1,10.1c-5.6,0-10.1-4.6-10.1-10.1c0-5.6,4.5-10.1,10.1-10.1C84.2,46.7,88.7,51.3,88.7,56.8z};
    }
}
\newcommand\orcidicon[1]{\href{https://orcid.org/#1}{\mbox{\scalerel*{
                \begin{tikzpicture}[yscale=-1,transform shape]
                \pic{orcidlogo};
                \end{tikzpicture}
            }{|}}}}
\usepackage[implicit=false]{hyperref}
\hypersetup{hidelinks,
	colorlinks=true,
	allcolors=black,
	pdfstartview=Fit,
	breaklinks=true}

\newcommand{\ie}{\textit{i}.\textit{e}.}

\usepackage{xspace}
\newcommand{\name}{Boost-MT\xspace}
\begin{document}

\title{Boosting Meta-Training with Base Class\\ Information for Few-Shot Learning}

\author{Weihao Jiang,  
Guodong Liu, 
Di He,  
Kun He${\textsuperscript{\orcidicon{0000-0001-7627-4604}}}$,~\IEEEmembership{Senior Member,~IEEE}
\thanks{
This paper is supported by National Natural Science Foundation of China (U22B2017). 
}
\thanks{}}

\markboth{}%
{Jiang \MakeLowercase{\textit{et al.}}: Boosting Meta-Training with Base Class Information for Few-Shot Learning}


\maketitle

\begin{abstract}
Few-shot learning, a challenging task in machine learning, aims to learn a classifier adaptable to recognize new, unseen classes with limited labeled examples. Meta-learning has emerged as a prominent framework for few-shot learning. Its training framework is originally a task-level learning method, such as Model-Agnostic Meta-Learning (MAML) and Prototypical Networks. And a recently proposed training paradigm called Meta-Baseline, which consists of sequential pre-training and meta-training stages, gains state-of-the-art performance. However, as a non-end-to-end training method, indicating the meta-training stage can only begin after the completion of pre-training, Meta-Baseline suffers from  higher training cost 
and suboptimal performance due to the inherent conflicts of the two training stages.
To address these limitations, we propose an end-to-end training paradigm consisting of two alternative loops. In the outer loop, we calculate cross entropy loss on the entire  training set while updating only the final linear layer. In the inner loop, we employ the original meta-learning training mode to calculate the loss and incorporate gradients from the outer loss  to guide the parameter updates. 
This training paradigm not only converges quickly but also outperforms existing  baselines, indicating that information from the overall training set and the meta-learning training paradigm could mutually reinforce one another. Moreover, being model-agnostic,  
our framework achieves significant performance gains, surpassing the baseline systems by approximate $1\%$.
\end{abstract}


\section{Introduction}
\label{section_1}
Deep learning has achieved remarkable success in the fields of text, image, and audio classification in recent years. Training these deep-learning models often requires a huge amount of data. However, when the available data is limited, the models often perform poorly due to the overfitting issue. This leads to the necessity of solving the few-shot learning problem \cite{siamese,one-shot}: how can we train a model to achieve good results using a small amount of data when facing new tasks? Meta-learning \cite{RelationNetwork,meta-reg,meta-proto} has recently been the most common framework to address such a few-shot learning problem. 

There are two stages in meta-learning, namely meta-training and meta-testing, with separate data used. 
Meta-training samples tasks from the whole training set. 
There are a support set and a query set for each task. 
The support set contains $N$ classes each with $K$ samples, formulated as the $N$-way $K$-shot problem, while the query set contains $N$ classes each with $Q$ samples. Meta-learning aims to train a robust model during the meta-training phase, which can effectively classify the query set with only $N \times K$ support samples in new tasks during meta-testing. 
 
Previous training methods for meta-learning roughly fall into two categories, \ie, optimization-based and metric-based, both of which are versions without pre-training. A good initialization parameter is learned during the optimization-based meta-training methods, which can be quickly adapted to new tasks in the meta-testing stage. The goal of metric-based methods is to learn a good representation embedding in the meta-training stage. In the meta-testing stage, the support set and query set are mapped to the same high-dimensional embedding space, and the classification is carried out by measuring the distance between the two sets. 

Meta-learning with pre-trained models has emerged in recent years, which includes two stages. In the pre-training stage, the whole training classes of data are used to train the feature extractor through supervised or unsupervised methods. In the meta-learning stage, the original meta-learning training method (such as Prototypical Networks\cite{PrototypicalNetwork}) is used to retrain the feature extractor. A representative work of this category is Meta-Baseline \cite{Meta-Baseline}, which outperforms the primary meta-learning methods and achieves new state-of-the-art performance.

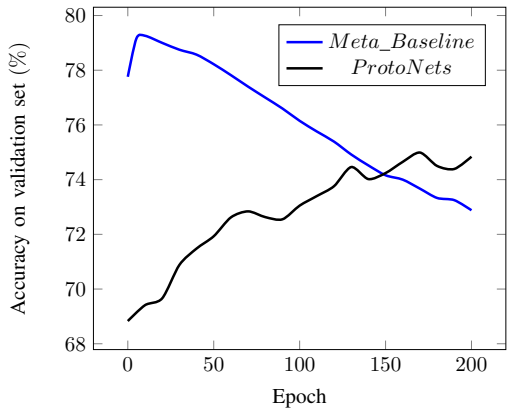
\begin{figure}[t] 
\centering 

\begin{tikzpicture}[scale=0.8] 

\begin{axis}[
    xlabel=Epoch, 
    ylabel=Accuracy on validation set $(\%)$, 
    tick align=inside, 
    legend style={at={(0.70,0.95))},anchor=north} 
    ]

\addplot[smooth,blue, line width = 1.2pt] plot coordinates { 
    (0,77.76)
    (5,79.16)
    (10,79.26)
    (20,79.00)
    (30,78.75)
    (40,78.57)
    (50,78.22)
    (60,77.82)
    (70,77.40)
    (80,77.00)
    (90,76.60)
    (100,76.15)
    (110,75.76)
    (120,75.39)
    (130,74.92)
    (140,74.52)
    (150,74.16)
    (160,74.00)
    (170,73.67)
    (180,73.33)
    (190,73.25)
    (200,72.88)

};
\addlegendentry{$Meta\_Baseline$}



    

\addplot[smooth,black, line width = 1.2pt] plot coordinates { 
    (0,68.83)
    (10,69.41)
    (20,69.66)
    (30,70.88)
    (40,71.48)
    (50,71.93)
    (60,72.62)
    (70,72.84)
    (80,72.63)
    (90,72.55)
    (100,73.05)
    (110,73.40)
    (120,73.76)
    (130,74.46)
    (140,74.02)
    (150,74.24)
    (160,74.65)
    (170,74.99)
    (180,74.50)
    (190,74.39)
    (200,74.84)

};
\addlegendentry{$ProtoNets$}
\end{axis}
\end{tikzpicture}
\caption{Accuracy of the second stage of Meta-Baseline 
on the validation dataset of $mini$ImageNet along with 5-way 5-shot training process. We also re-implement primary Protypical Networks ($ProtoNets$) under the same setting.}
\label{fig:dif-measure}
\end{figure}

However, as shown in Figure \ref{fig:dif-measure}, we observe that in the second stage of Meta-Baseline, the performance on the validation set first increases, then decreases, and finally, is even lower than that of the Prototypical Networks (ProtoNets), indicating that the training of the first and second stages have some negative impact with each other. 
Moreover, we have tried to change the consine metric of Meta-Baseline to other metrics, including Euclidean distance, cosine similarity plus Euclidean distance, Manhattan distance and Chebyshev distance, but we still could not mitigate the issue,
indicating the difficulty of utilizing the representation information of pre-training in the second stage of Meta-Baseline.

\IEEEpubidadjcol

Besides, 
Meta-Baseline
is not an end-to-end training method, which means the second training stage must wait until the first stage converges, resulting in a significantly longer overall training time. Moreover, the accuracy of the first and second stages is not proportional, indicating that the model with higher accuracy in the first stage may not perform better in the second stage. Therefore, multiple models must be trained in the second stage to achieve better results. In summary, simply concatenating the two stages of training is not helpful. 

In this work, 
we design a new end-to-end training approach to boost meta-training with base class information, termed \name. 
In this method, there are two loops that are executed alternately.
In the outer loop, we calculate the classification loss of one large batch from the whole training set and only update the final linear layer. In the inner loop, we use the meta-learning method to calculate the loss of several divided episodic tasks and update the model by incorporating inner loss and outer loss. 
Our method not only can combine pre-training and meta-training into a single end-to-end model but also can quickly converge and achieve competitive results,  which is more consistent with the deep learning paradigm. Additionally, our approach can avoid mutual subtraction 
between the two stages of Meta-Baseline. 

In addition, our model is very simple and easy to use. There is almost no increase in the amount of parameters compared with the previous model. 
The only extra hyper-parameter is the proportion of outer and inner loops.
We further conduct ablation experiments to explore the effectiveness of our model. 
Our main contributions are summarized as follows:

$\bullet$ We propose a new few-shot learning framework, that is end-to-end and can converge quickly, providing new motivations for future research on few-shot learning.

$\bullet$ To the best of our knowledge, this work is the first to adopt gradient information from base class 
to the training process of few-shot learning. In this way, 
the pre-training and meta-learning can guide each other, thus avoiding the mutual subtraction between the two stages observed in existing methods. 

$\bullet$ Our evaluation on two benchmark datasets, $mini$ImageNet and $tiered$ImageNet datasets, demonstrate the competitive performance of our method compared to the state-of-the-art baselines.

$\bullet$ Our approach is model-agnostic. The incorporation of our framework into CAN~\cite{can} and FEAT~\cite{FEAT} results in a notable performance enhancement, yielding approximate $1\%$ improvement over the baseline systems.

\section{Related Work}
This section reviews some popular methods on few-shot learning, which fall into the following three categories.  

\subsection{Meta-learning for few-shot learning}
Most of the existing few-shot learning methods rely on the meta-learning framework, which can be roughly divided into two categories, 
\ie, 
optimization-based~\cite{MAML,meta-sgd,First-Order,optim-MAML,ANIL,BOIL,LEO} and metric-based~\cite{Revisit,TADAM,RelationNetwork,MatchingNetwork,PrototypicalNetwork,TPN,deepemd}. 

The optimization-based methods aim to quickly learn the model parameters that can be optimized when encountering new tasks. The representative work of this category 
is MAML~\cite{MAML}, which obtains good initialization parameters through internal and external training in the meta-training stage. The model with good parameters can obtain good results with a few steps of update in the meta-testing stage. MAML has inspired many follow-up efforts, such as ANIL~\cite{ANIL}, BOIL~\cite{BOIL}, LEO~\cite{LEO} and Con-MetaReg~\cite{meta-reg}.

Metric-based methods embed support images and query images into the same space and classify query images by calculating the distance or similarity. ProtoNets~\cite{PrototypicalNetwork} compute a prototype for each class and classify query images by calculating the Euclidean distance. Relation Networks~\cite{RelationNetwork} calculate distances between support images and query images via a relation module. TADAM~\cite{TADAM} boosts ProtoNets by metric scaling and metric task conditioning. DeepEMD~\cite{deepemd} employs the Earth Mover’s Distance (EMD) as a metric to compute a structural distance between dense support images and query images.

\subsection{Transfer learning for few-shot learning}
The transfer learning framework for few shot learning~\cite{dynamic,baseline++,goodembedd,a-baseline,partial-transfer} uses a simple 
scheme to train a classification model on the overall training 
set. During testing, 
it removes the classification head and retains the feature extraction part, then trains a new classifier based on the support set from testing data. 
This approach yields results comparable to 
meta-learning. 

Specifically, Dynamic Classifier~\cite{dynamic} trains a feature extractor and a small sample category weight generator using all the base classes. In the testing stage, the weight generator can generate weight vectors for each level class with a few samples. 
Baseline++~\cite{baseline++} replaces the linear classifier with a cosine classifier in the training stage, and directly trains the feature extractor and cosine classifier using all the base classes. During testing, it fixes the feature extractor and trains a new classifier with limited labeled examples in novel classes, 
and achieved good results. 
Good-Embed~\cite{goodembedd} uses ordinary classification or self-supervised learning to train a feature extractor in the training stage. Then it fits a linear classifier during testing on features extracted by the pre-trained network for each task, and further applies self-distillation to the pre-trained network.

\subsection{Meta-learning with pre-training}
Inspired by the transfer learning for few-shot learning methods, several recent studies~\cite{Meta-Baseline,mpm,deepbdc,P>M>F,with-contrastive,FEAT,FRN} propose to use a meta-learning model with pre-training. Meta-Baseline~\cite{Meta-Baseline} is the first of this kind, 
which explores the advantages of the overall classification model and the meta-learning model, and proposes a baseline method that continues the meta-learning in the converged classifier by using the evaluation measure of cosine nearest centroid. 

Many subsequent few-shot learning methods follow the paradigm of Meta-Baseline. 
For instance, Meta DeepBDC~\cite{deepbdc}, based on a typical transfer learning framework of Good-Embed, uses plenty of annotation data to train a better basic model that obtains the embedded features of the image and then uses the Brownian distance covariance measurement for meta-learning training based on the framework of ProtoNets.
Contrastive-FSL~\cite{with-contrastive} proposes
a novel contrastive learning-based framework that seamlessly integrates contrastive learning into both pre-training and meta-training.
P$>$M$>$F~\cite{P>M>F} uses an unsupervised training method to obtain feature extractors in pre-training, then uses meta-learning to optimize the model. It finally fine-tunes the model using sample of the support set in the novel class, pushing the application of this method to the limit.

Although the meta-learning method with pre-training has achieved good results, it is not an end-to-end method that needs higher training cost and does not conform to the deep learning paradigm. 
In this work, we propose a novel end-to-end training framework with outer loop and inner loop, that leverages information from the base class to guide meta-learning instead of relying on direct pre-training. 

\section{Preliminary}
We first establish preliminaries on few-shot learning in Section \ref{section_3_1}, then we briefly introduce the flow of the Meta-Baseline method in Section \ref{section_3_2} and the objective discrepancy in meta-learning in Section \ref{section_3_3}.

\subsection{Notations}
\label{section_3_1}
Given a labeled dataset $\mathcal{D}^{base}$ with a large number of images, the goal of few-shot learning is to learn concepts in novel classes $\mathcal{D}^{novel}$. Note that $\mathcal{D}^{base}$ and $\mathcal{D}^{novel}$ are not intersected.
In the meta-training stage, the few-shot learning problem randomly samples many tasks $\mathcal{T} = {\{ (\mathcal{D}_t^{spt}, \mathcal{D}_t^{qry}) 
\}}_{t=1}^T$ from $\mathcal{D}^{base}$.
The support set $\mathcal{D}_t^{spt} = {\{(\mathbf{x}_i^t, y_i^t) \}}_{i=1}^{NK}$ and the query set $\mathcal{D}_t^{qry} = {\{(\mathbf{x}_j^t, y_j^t) \}}_{j=1}^{NQ}$ in each task are sampled under the $N$-way $K$-shot setting (support set consists of $N$ classes each with $K$ images and query set consists of $N$ classes each with $Q$ images). $S_n$  denotes the set of examples labeled with class $n$ in each task.

\subsection{Meta-Baseline}
\label{section_3_2}
Meta-Baseline includes two training stages. 
The first stage is for the pre-training. 
A classification model is trained on $\mathcal{D}^{base}$. The model is composed of feature extractor $f_\theta$ and classifier $fc_\omega$.
The second stage is meta-learning training, which calculates the distance between features using metric learning. 
First, Meta-Baseline sends the support set into the feature extractor to obtain each sample feature and averages all sample features of each class to get the prototype of each class $n$:
\begin{align}\label{e1}
    \mathbf{c}_n = \frac{1}{|S_n|} \sum_{(\mathbf{x},y) \in {S_n}}f_{\theta}(\mathbf{x}).
\end{align}
Similarly, a query point $\mathbf{x}$ is sent to the feature extractor to obtain its feature. The cosine similarity between $\mathbf{x}$ and the support set prototypes is calculated to obtain the probability that $\mathbf{x}$ belongs to class $n$:
\begin{align}\label{e2}
    p_{\theta}(y=n|\mathbf{x}) = \frac{exp(d(f_{\theta}(\mathbf{x}),\mathbf{c}_n ))}{\sum_{n=1}^{N}exp(d(f_{\theta}(\mathbf{x}),\mathbf{c}_n))},
\end{align}
where $d(\cdot,\cdot)$ denotes the cosine similarity between two vectors.
Finally, the cross entropy $\mathcal{L}$ is used as the loss function.

Meta-Baseline is meta-learning over a converged feature extractor on its evaluation metric, which means the
meta-training stage can only begin after the completion of the pre-training. In addition, further meta-learning based on the convergent model parameters can inherit better class transfer ability, but it also limits the model to find better convergence points.

\subsection{Objective discrepancy in meta-learning}
\label{section_3_3}
Despite the improvements of meta-learning with pre-training, we observe that the performance of the meta-learning stage is prone to overfitting.
To better illustrate this issue, we introduce two concepts proposed in the Mate-Baseline~\cite{Meta-Baseline}, namely base class generalization and novel class generalization. Base class generalization is measured by sampling tasks from images that are not visible from the base classes, while novel class generalization refers to the performance of a few-shot task sampled from novel classes. Previous studies~\cite{Meta-Baseline} have found that in the meta-learning stage, as the base class generalization increases, the novel class generalization actually decreases. 
This observation suggests that over a converged conventional classification model, the meta-learning objective itself, \ie, making the embedding generalize better in few-shot tasks from base classes, can hurt the performance of few-shot tasks from novel classes.
Some researchers~\cite{Meta-Baseline} speculate that in some cases, optimizing towards training goals in a consistent form as a testing objective may have even negative impact. It is also possible that the whole-classification learning embedding has stronger class transferability, while meta-learning makes the model perform better in $N$-way $K$-shot tasks, but tends to lose the class transferability.

To address this issue, we integrate conventional classification optimization at each step with few-shot task classification optimization to circumvent the staged optimization process.
In our method, the outer loop iterates on the batch of base class, and the inner loop iterates on the few-shot tasks. 
The model of the outer loop consists of a feature extractor and classifier, while the model of the inner loop consists of a feature extractor and cosine metric.
The outer loop updates the classifier via classification loss on the batch of base class to achieve class transferability but keeps the feature extractor unchanged.  
The inner loop updates the feature extractor by the gradient of the current few-shot task loss, minus a bias: the gradient of the current few-shot task on outer weight minus the batch gradient of outer weight on classification loss.  

In summary, we utilize the gradient of the classification loss to update few-shot learning within the context of meta-learning, enabling the model to achieve superior generalization while adapting to the few-shot learning task.

\section{Methodology}
In this section, we first provide our motivation 
and then introduce our Boost-MT method. 

\subsection{Motivation}
\label{section_4_1}
Many prior works~\cite{goodembedd, Meta-Baseline} have demonstrated that few-shot classifiers benefit significantly from non-episodic pre-training. Therefore, when the meta-training is continued based on the pre-training, the model performance should continue to improve until converging stably.  
However, as shown in Figure \ref{fig:dif-measure}, we observe that in the second stage of Meta-Baseline, the performance on the validation set decays quickly after a short increase, 
and finally, is even lower than that of the Prototypical Networks.
Such phenomenon indicates that the pre-training restricts the meta-learning to further improve performance on few-shot tasks.



\begin{figure}
\centering
\subfloat[On $mini$Imagenet]{
\begin{tikzpicture}[scale=0.48] 

          \begin{axis}[
            xlabel=Epoch, 
            ylabel=Accuracy $(\%)$, 
            tick align=inside, 
            legend style={at={(0.75,0.97))},anchor=north} 
            ]
        
        
        
          \addplot[smooth,blue, line width = 1.5pt] plot coordinates { 
            (0,77.76)
            (5,79.16)
            (10,79.26)
            (20,79.00)
            (30,78.75)
            (40,78.57)
            (50,78.22)
            (60,77.82)
            (70,77.40)
            (80,77.00)
            (90,76.60)
            (100,76.15)
        
        };
          \addlegendentry{Few-shot}
        \addplot[dashed,smooth,blue,line width = 1.5pt] plot coordinates { 
            (0,80.48)
            (5,81.38)
            (10,81.50)
            (15,81.58)
            (20,81.31)
            (25,81.30)
            (30,80.79)
            (40,80.73)
            (50,80.07)
            (60,79.35)
            (70,78.89)
            (80,77.77)
            (90,77.55)
            (100,77.12)
        
        };
          \addlegendentry{Conventional}
          \end{axis}
\end{tikzpicture}
\label{fig:sub1}}
\hfil
\subfloat[On $tiered$Imagenet]{
\begin{tikzpicture}[scale=0.48] 

      \begin{axis}[
            xlabel=Epoch, 
            ylabel=Accuracy $(\%)$, 
            tick align=inside, 
            legend style={at={(0.75,0.97))},anchor=north} 
            ]
        
        \addplot[smooth,blue, line width = 1.5pt] plot coordinates { 
            (0,83.34)
            (5,82.09)
            (10,80.62)
            (20,78.09)
            (30,76.96)
            (40,75.75)
            (50,75.03)
            (60,74.03)
            (70,72.76)
            (80,72.81)
            (90,72.22)
            (100,71.54)
        
        };
          \addlegendentry{Few-shot}
        \addplot[dashed,smooth,blue,line width = 1.5pt] plot coordinates { 
            (0,71.78)
            (5,65.40)
            (10,63.03)
            (20,62.97)
            (30,62.80)
            (40,62.55)
            (50,62.13)
            (60,62.57)
            (70,62.53)
            (80,62.58)
            (90,62.91)
            (100,62.81)
        };
          \addlegendentry{Conventional} 
      \end{axis}
\end{tikzpicture} 
\label{fig:sub2}}
\caption{
Few-shot as well as conventional classification accuracy at the second stage of Meta-Baseline along with the training process. 
}
\label{fig:conventional}
\end{figure}
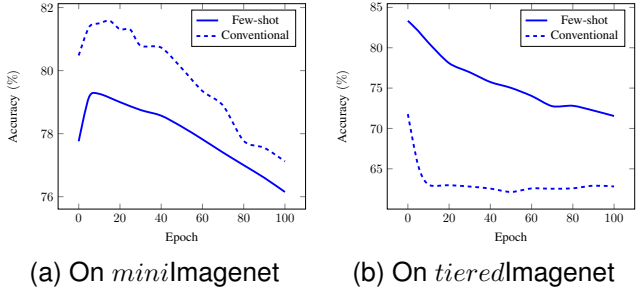

To investigate the underlying factors behind this phenomenon, we conduct 
more experiments on two datasets. 
To check whether the feature extractor
of the second stage is still good for conventional classification task, we freeze the feature extractor and fine-tune the classification headers obtained in the first stage to execute a conventional classification task. 
The results are illustrated in Figure \ref{fig:conventional}. We observe that during the second stage,
performance of the conventional classification task shows similar trends as that of the few-shot task. 
Such decays indicate that second-stage meta-learning adversely affects the model's class transferability.

Building upon this motivation, we posit that by incorporating both tasks into consideration during the feature extractor's updates, the model can attain the capability to address both tasks simultaneously, having higher generalization capability and hence be more robust to the few-shot learning task on new classes. 
Thereby, we propose a novel framework termed Boost-MT that leverages information from the base class to guide meta-learning instead of relying on direct pre-training. 

\subsection{Boosting meta-training with base class information}
\label{section_3_4}
\begin{figure*}
\begin{center}
  \includegraphics[width=0.9\linewidth]{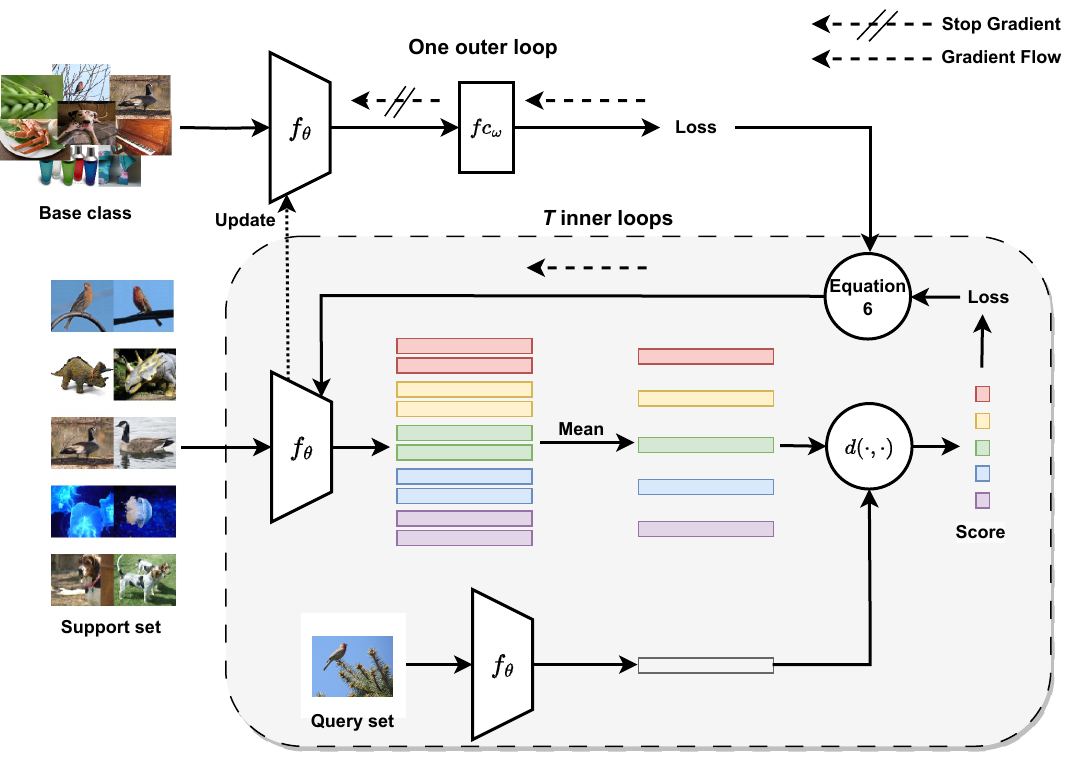} 
\end{center}
   \caption{The main framework of our model. In the outer loop, we calculate the classification loss of one large batch from base classes and only update the linear classifier. In the inner loop, we use the meta-learning method to calculate the loss of tasks and update the model by inner loss and outer loss. 
   The two loops are executed alternately, with $T$ inner loops per one outer loop. }
\label{fig:framwork}
\end{figure*}

As shown in Figure \ref{fig:framwork}, our method consists of 
two loops. The outer loop is executed $S$ times in one epoch, and each outer loop corresponds to $T$ inner loops. The network model consists of a feature extractor $f_\theta$ used in both the outer loop and inner loop, a classification header $fc_\omega$ used in the outer loop, and a cosine similarity measurement function $d(\cdot,\cdot)$ used in the inner loop. Here we randomly initialize $f_\theta$ and $fc_\omega$.

In each outer loop, we first randomly sample a large set $\mathcal{D}^{out}$ with $N_b$ images from $D^{base}$. Then we calculate the loss $\Tilde{\mu}$ of $D^{out}$ by \eqref{e3} and only update $fc_{\omega}$. 
When all outer loops in one epoch are done, all samples in  $\mathcal{D}^{base}$ are traversed once.
Before the first inner loop begins, we randomly sample $T$ tasks $\mathcal{T} = {\{ (\mathcal{D}_t^{spt}, \mathcal{D}_t^{qry}) \}}_{t=1}^T$ from $\mathcal{D}^{base}$ as standard few-shot learning.

\begin{align}\label{e3}
    \Tilde{\mu} = \frac{1}{N_b}\sum_{(\mathbf{x},y) \in \mathcal{D}^{out}}\mathcal{L}_{\mathbf{x}}(f_{\Tilde{\theta}_{s-1}},fc_{\Tilde{\omega}_{s-1}}),
\end{align}
where $s=\left\{1,...,S\right\}$, and $S$ is the number of outer loops. $\Tilde{\theta}_{s}$ and $\Tilde{\omega}_{s}$ represent the parameters of $f_{\theta}$ and $fc_{\omega}$ in the $s$-th iteration of outer loop, respectively.

In each inner loop, one task is calculated at a time. The feature extractor $f_{\theta}$ extracts the embedded representations of the support set and query set. We calculate the prototype of each class in the support set by \eqref{e1}. 
For each sample in the query set, we use cosine similarity measurement $d(\cdot,\cdot)$ and predict its label by \eqref{e2}.
The loss of query point $\mathbf{x}$ is calculated by the cross entropy loss.
The inner loss of the query set $\sigma$ is calculated by $f_{\theta}$ with inner loop parameters and cosine similarity measurement $d(\cdot,\cdot)$, as in \eqref{e4}.
The outer loss of the query set $\Tilde{\sigma}$ is calculated by $f_{\theta}$ with outer loop parameters and $d(\cdot,\cdot)$, as in \eqref{e5}.
The inner and outer losses of the query set and the outer loss of $\mathcal{D}^{out}$ will both be used to update the parameters of $f_{\theta}$, as in \eqref{e6}.
In the next inner loop, the updated parameters in the 
previous inner loop are used to calculate inner loss until all inner loops are done.
Then, the parameters of $f_\theta$ in the next outer loop are inherited from the previous inner loop.
The details are shown in Algorithm \ref{alo1}.

\begin{align}\label{e4}
    \sigma = \frac{1}{NQ}\sum_{(\mathbf{x},y) \in \mathcal{D}_t^{qry}}\mathcal{L}_{\mathbf{x}}(f_{\theta_{t-1}},d(\cdot,\cdot)),
\end{align}
where $t=\left\{1,...,T\right\}$, and $T$ is the number of inner loops. ${\theta}_{t}$ represents the parameters of $f_{\theta}$ in the $t$-th iteration of inner loop.
\begin{align}\label{e5}
    \Tilde{\sigma} = \frac{1}{NQ}\sum_{(\mathbf{x},y) \in \mathcal{D}_t^{qry}}\mathcal{L}_{\mathbf{x}}(f_{\Tilde{\theta}_{s-1}},d(\cdot,\cdot)),
\end{align}
\begin{align}\label{e6}
    \theta_t = \theta_{t-1} - \beta(\nabla\sigma - \nabla\Tilde{\sigma} + \nabla\Tilde{\mu}) .
\end{align}

As analyzed in Section \ref{section_1}, a fixed embedding space from pre-training is difficult to exploit for meta-training.
Unlike the meta-learning method with pre-training, although we calculate the loss on the base classes in the outer loop, we do not update the parameters of the feature extractor, avoiding the direct update of representation information by $\mathcal{D}^{base}$. 

The key aspect of our method is that the loss function is updated using the loss in each task during the inner loop of updating the feature extractor, which simultaneously draws the feature extractor in the outer loop and the last update.
Instead of continuing the meta-learning on the already convergent classification feature extractor, we use the gradient generated by classification training to correct the optimization direction of the meta-learning, to guide each meta-training, which not only can inherit the class transfer ability brought by classification training but also can better adapt to the task of meta-learning.
Therefore, our end-to-end method can still utilize the information from $\mathcal{D}^{base}$ by updating inner parameters through outer loss and inner loss, which avoids mutual subtraction between the two stages and boosts meta-training.
\begin{figure}
\centering
\subfloat[]{\includegraphics[width=0.45\linewidth]{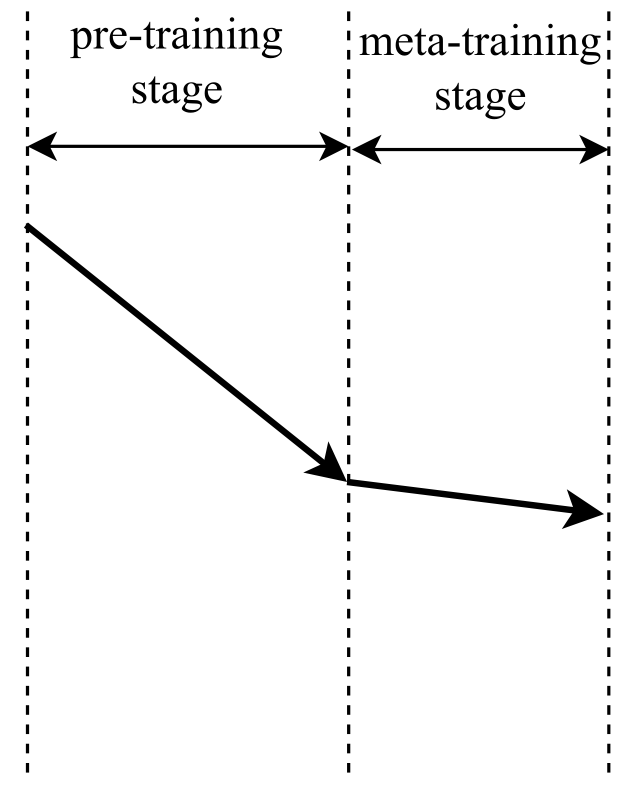}%
\label{fig1}}
\hfil
\subfloat[]{\includegraphics[width=0.475\linewidth]{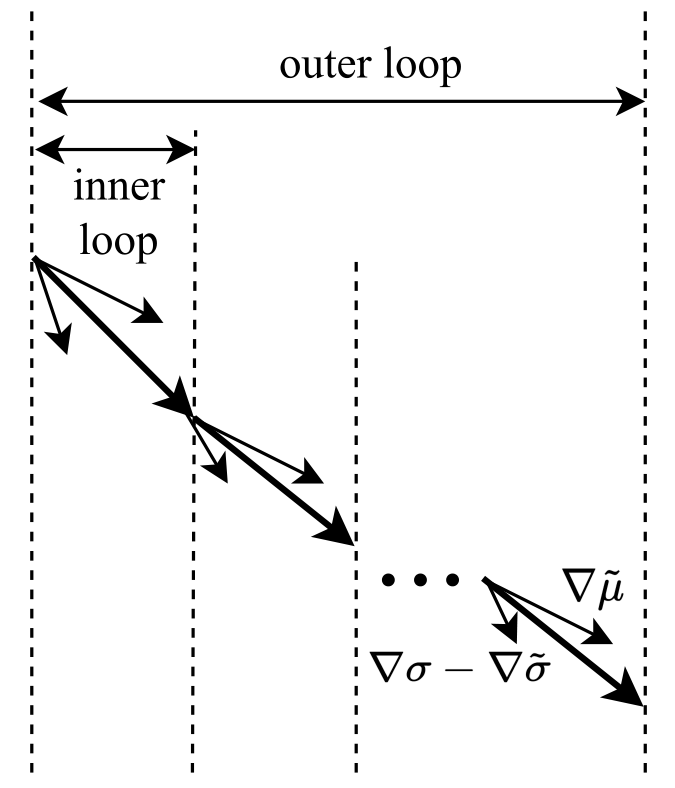}%
\label{fig2}}
\caption{The parameter update procedures of the Meta-Baseline and our model.  In Meta-Baseline, meta-training continues on the model parameters that have been converged in pre-training, as shown in (a). 
In contrast, our model employs an end-to-end training process as illustrated in (b). In the outer loop, we compute the classification loss for a large batch from base classes. In the inner loop, we utilize the meta-learning method to calculate task losses and update the model based on both inner and outer losses.}
\label{fig:svrg_metabaseline}
\end{figure}

\begin{algorithm}[t]
\caption{The Boost-MT Method}\label{alg:alo1}
\begin{algorithmic}[1]
\STATE \textbf{Input:} 
base classes $\mathcal{D}^{base}$,
feature extractor with parameters $\theta$: $f_{\theta}$,
classification header with parameters $\omega$: $fc_{\omega}$,
hyperparameters on step size: $\alpha, \beta$ 
\STATE \textbf{Output:} $\theta$
\STATE \textbf{Require:} RandSample($\mathcal{D},N$): randomly sample $N$ images from $\mathcal{D}$ 
\STATE Initialize $\Tilde{\theta}_0, \Tilde{\omega}_0$
\FOR{$s=1,2,3, \dots,S$}
    \STATE $\mathcal{D}^{out} \leftarrow$ RandSample$(\mathcal{D}^{base}, N_b)$
    \STATE Compute the loss of $\mathcal{D}^{out}$ by \eqref{e3}
    \STATE $\Tilde{w}_s = \Tilde{w}_{s-1} - \alpha\nabla\Tilde{\mu}$
    \STATE $\theta_{0}=\Tilde{\theta}_{s-1}$
    \STATE Sample $\mathcal{T} = {\{ (\mathcal{D}_t^{spt}, \mathcal{D}_t^{qry}) 
    \}}_{t=1}^T$ from $\mathcal{D}^{base}$
    \FOR{$t=1,2,3,\dots,T$}
        \STATE Compute the prototype of the $t$ task by \eqref{e1}
        \STATE Compute the probability of the query point $\mathbf{x}$ by \eqref{e2}
        \STATE Compute the inner loss of $\mathcal{D}_t^{qry}$ with inner loop parameters by \eqref{e4}
        \STATE Compute the outer loss of $\mathcal{D}_t^{qry}$ with outer loop parameters by \eqref{e5}
        \STATE Update $\theta$ based on \eqref{e6}
    \ENDFOR
    \STATE $ \Tilde{\theta}_s = \theta_T$
\ENDFOR
\STATE $ \theta = \Tilde{\theta}_S$
\STATE \textbf{return}  $\theta$
\end{algorithmic}
\label{alo1}
\end{algorithm}

\subsection{Difference to meta-learning with pre-training}
The meta-learning with the pre-training method typically involves two distinct stages, as illustrated in Figure \ref{fig1}, namely pre-training and meta-training. During the pre-training stage, the model undergoes conventional classification training to acquire convergent feature extractor parameters. Subsequently, in the meta-training stage, the model is retrained using a few-shot learning mechanism, leveraging the previously obtained convergent model parameters.
However, employing further optimization within the already convergent model space poses the risk of encountering overfitting issues, as depicted in Figure \ref{fig:dif-measure}. This scenario makes it challenging to achieve additional enhancements in model accuracy. 

In contrast, our proposed approach deviates from this phased training paradigm. Instead, it adopts a strategy that combines classification training with several few-shot learning tasks at each iteration, guiding the parameter updates, as depicted in Figure \ref{fig2}.
To elaborate, the outer loop initially calculates the classification loss for a batch of data, while the inner loop computes the loss for the few-shot task. The model parameters are then updated based on both losses 
as in \eqref{e6}. By fully updating the model at each step, our approach mitigates the risk of premature overfitting, ensuring a more robust training process.

\section{Experiments}
In this section, we first describe the experimental results, comparing the performance of our method against various baselines on two common benchmark datasets. Secondly, we conduct a detailed comparison with the current strongest method of Meta-Bseline. Then, we do an ablation study to show the effectiveness of our method and a parameter study to show the impact of the number of inner loops per outer loop. In the end, we discuss the possible extension of our method.

\subsection{Datasets}
In the standard few-shot classification task, commonly used datasets are Omniglot~\cite{one-shot}, CUB-200-2011 (Birds)~\cite{cub}, $mini$ImageNet~\cite{MiniImageNet}, and $tiered$ImageNet~\cite{tieredImageNet}. Omniglot is used more often in early studies. 
Many later algorithms have achieved very high accuracy in training and testing on this dataset. The scale of the CUB dataset is between $mini$ImageNet and $tiered$ImageNet, so we use $mini$ImageNet and $tiered$ImageNet for few-shot classification, which are also the two most commonly used datasets.
We also conduct cross-domain experiments by training on $mini$ImageNet and testing on CUB.

The $mini$ImageNet dataset was created by the authors of Matching Network~\cite{MatchingNetwork} and is currently the most popular benchmark. It contains 100 categories sampled from ILSVRC-2012~\cite{ImageNet}. Each category contains 600 images, and the size of each image is 84 × 84, with 60,000 images in total. Under standard settings, it is randomly 
partitioned into training, validation, and testing sets, each containing 64, 16, and 20 categories.

The $tiered$ImageNet dataset is another commonly used benchmark. Although it is also from ILSVRC-2012, compared to the former, 
it has a more extensive data scale. There are 34 super-categories, divided into training, validation, and testing sets, with 20, 6, and 8 super-categories, respectively. It contains 608 categories, corresponding to 351, 97, and 160 categories in each partitioned dataset. This dataset is more challenging because it is more difficult for the model to identify samples of different categories from the same super-categories. The base classes and novel classes come from different super-categories. 

The CUB dataset contains 200 classes and 11,788 images in total. Following the protocol of Hilliard et al. (2018)~\cite{cub}, we randomly split the dataset into 100 base, 50 validation, and 50 novel classes.


\subsection{Implementation details}
In order to keep consistent with the Meta-Baseline, we use the ResNet-12 network backbone with a feature dimension of 512. For the outer loop, we use the SGD optimizer with momentum 0.9 to update the parameters in the linear layer, the learning rate starts from 0.1, and the decay factor is set to 0.1. For the inner loop, we set 10 tasks to update the feature extractor. On $mini$ImageNet, we train 100 epochs with batch size 128, and the learning rate decays at epoch 30 and 60. On $tiered$ImageNet, we train 100 epochs with batch size 256, and the learning rate decays at epoch 30 and 60. As in the previous work, we use the validation set to select models. We also use commonly standard data augmentation methods, including random resized loop and horizontal flip. However, we do not set the cosine scaling parameter $\tau$.
During the meta-testing, we randomly sample 1500 tasks with 15 query images per class and report the mean accuracy together with the corresponding $95\%$ confidence interval.

\begin{table}[t]
\caption{Comparison to prior works on $mini$ImageNet. Average 5-way accuracy with 95$\%$ confidence interval.}
\label{tab:miniImage}
\begin{center}
\scalebox{0.9}{
    \begin{tabular}{|l|c|c|c|}
    \hline
    \bf Model & \bf Backnone & \bf 1-shot & \bf   5-shot \\
    \hline
    Baseline++~\cite{baseline++} & ResNet-18 & 51.87 ± 0.77 & 75.68 ± 0.63 \\
    MataOptNet~\cite{metaoptnet} & ResNet-12 & 62.64 ± 0.61 & 78.63 ± 0.46 \\
    Shot-Free~\cite{shot-free} & ResNet-12 & 59.04 ± 0.43 & 77.64 ± 0.39 \\
    TADAM~\cite{TADAM} & ResNet-12 & 58.50 ± 0.30 & 76.70 ± 0.30\\
    MTL~\cite{MTL} & ResNet-12 & 61.20 ± 1.80 &	75.50 ± 0.80\\
    SLA-AG~\cite{sla-ag} & ResNet-12 & 62.93 ± 0.63 & 79.63 ± 0.47\\
    MPM~\cite{mpm} & ResNet-12 & 62.79  & 78.51 \\
    ProtoNets + TRAML~\cite{boosting} & ResNet-12 & 60.31 ± 0.48 & 77.94 ± 0.57\\
    ProtoNets~\cite{PrototypicalNetwork} & ResNet-12 & 60.37 ± 0.83 & 78.02 ± 0.57\\ 
    Classifier-Baseline~\cite{Meta-Baseline}  & ResNet-12 & 58.91 ± 0.23 & 77.76 ± 0.17\\
    Meta-Baseline~\cite{Meta-Baseline}  & ResNet-12 & 63.17 ± 0.23 & 79.26 ± 0.17\\ 
    \hline
    Boost-MT (ours)  & ResNet-12 & \bf 64.01 ± 0.97 & \bf 81.00 ± 0.59\\
    \hline
    \end{tabular}
    }
\end{center}
\end{table}
\begin{table}[t]
\caption{Comparison to prior works on $tiered$ImageNet. Average 5-way accuracy with 95$\%$ confidence interval.}
\label{tab:tiered}
\begin{center}
\scalebox{0.9}{
\begin{tabular}{|l|c|c|c|}
    \hline
    \bf Model & \bf Backnone & \bf 1-shot & \bf 5-shot \\
    \hline
    LEO~\cite{LEO} & WRN-28-10 & 66.33 ± 0.05 & 81.44 ± 0.09\\
    MPM~\cite{mpm} & WRN-28-10 & 67.58  & 83.93 \\
    MetaOptNet~\cite{metaoptnet} & ResNet-12 & 65.99 ± 0.72 & 81.56 ± 0.53\\
    MTL~\cite{MTL} & ResNet-12 & 65.62 ± 1.80 & 80.61 ± 0.90\\
    AM3~\cite{AM3} & ResNet-12 & 67.23 ± 0.34 & 78.95 ± 0.22\\
    Shot-Free~\cite{shot-free} & ResNet-12 & 66.87 ± 0.43 & 82.64 ± 0.43\\
    DSN-MR~\cite{dsn-mr} & ResNet-12 & 67.39 ± 0.82 & 82.85 ± 0.56\\
    ProtoNets + Rotation~\cite{protonet-rotation} & ResNet-18 & – & 78.90 ± 0.70\\
    ProtoNets~\cite{PrototypicalNetwork} & ResNet-12 & 65.65 ± 0.92 & 83.40 ± 0.65\\ 
    Classifier-Baseline~\cite{Meta-Baseline} & ResNet-12 & 68.07 ± 0.26 & 83.74 ± 0.18\\
    Meta-Baseline~\cite{Meta-Baseline} & ResNet-12 & 68.62 ± 0.27 & 83.74 ± 0.18\\
    \hline
    Boost-MT (ours)  & ResNet-12 & \bf 69.73 ± 0.71 & \bf 84.91 ± 0.49\\
    \hline
    \end{tabular}}
\end{center}
\end{table}
\begin{figure*}
\centering

\subfloat[Meta-Baseline]{
\begin{tikzpicture}[scale=0.8] 

          \begin{axis}[
            xlabel=Epoch, 
            ylabel=Accuracy on validation set $(\%)$, 
            tick align=inside, 
            legend style={at={(0.80,0.52))},anchor=north} 
            ]
        
          \addplot[smooth,blue, line width = 1.2pt] plot coordinates { 
            (0,58.91)
            (5,62.73)
            (10,62.92)
            (15,62.93)
            (20,62.56)
            (25,62.32)
            (30,61.79)
            (40,61.71)
            (50,60.82)
            (60,60.76)
            (70,60.06)
            (80,59.00)
            (90,58.74)
            (100,58.32)
        
        };
          \addlegendentry{$mini$ 5-1}
        
          \addplot[smooth,green, line width = 1.2pt] plot coordinates { 
            (0,77.76)
            (5,79.16)
            (10,79.26)
            (20,79.00)
            (30,78.75)
            (40,78.57)
            (50,78.22)
            (60,77.82)
            (70,77.40)
            (80,77.00)
            (90,76.60)
            (100,76.15)
        
        };
          \addlegendentry{$mini$ 5-5}
        \addplot[smooth,red, line width = 1.2pt] plot coordinates { 
            (0,68.00)
            (5,68.07)
            (10,67.14)
            (20,65.12)
            (30,64.09)
            (40,63.25)
            (50,62.67)
            (60,62.12)
            (70,60.44)
            (80,61.07)
            (90,60.48)
            (100,60.39)
        
        };
          \addlegendentry{$tiered$ 5-1}
        \addplot[smooth,black, line width = 1.2pt] plot coordinates { 
            (0,83.34)
            (5,82.09)
            (10,80.62)
            (20,78.09)
            (30,76.96)
            (40,75.75)
            (50,75.03)
            (60,74.03)
            (70,72.76)
            (80,72.81)
            (90,72.22)
            (100,71.54)
        
        };
          \addlegendentry{$tiered$ 5-5}
          \end{axis}
\end{tikzpicture}
\label{fig:sub1}}
\hfil
\subfloat[Boost-MT]{
\begin{tikzpicture}[scale=0.8] 

      \begin{axis}[
            xlabel=Epoch, 
            ylabel=Accuracy on validation set $(\%)$, 
            tick align=inside, 
            legend style={at={(0.70,0.35))},anchor=north} 
            ]
        
      \addplot[smooth,blue, line width = 1.2pt] plot coordinates { 
            (0,30.68)
            (1,34.04)
            (5,45.53)
            (10,51.68)
            (15,54.89)
            (20,58.91)
            (25,59.41)
            (30,60.09)
            (35,62.17)
            (40,61.79)
            (45,61.78)
            (50,60.84)
            (55,60.72)
            (60,62.24)
            (65,60.05)
            (70,62.18)
            (75,60.91)
            (80,60.75)
            (85,61.29)
            (90,60.85)
            (95,61.97)
            (100,60.71)
        };
      \addlegendentry{$mini$ 5-1}
        
        
      \addplot[smooth,green, line width = 1.2pt] plot coordinates {
            (0,41.01)
            (5,63.61)
            (10,68.82)
            (15,73.32)
            (20,76.74)
            (25,76.42)
            (30,78.68)
            (35,78.84)
            (40,79.01)
            (45,79.23)
            (50,79.49)
            (55,79.44)
            (60,79.07)
            (65,78.04)
            (70,79.40)
            (75,80.08)
            (80,79.05)
            (85,78.65)
            (90,80.27)
            (95,79.04)
            (100,79.15)
        };
      \addlegendentry{$mini$ 5-5}
        
      \addplot[smooth,red, line width = 1.2pt] plot coordinates {
            (0,49.14)
            (5, 60.97)
            (10,64.62)
            (15,64.69)
            (20,64.87)
            (25,64.60)
            (30,65.66)
            (35,65.10)
            (40,66.19)
            (45,65.78)
            (50,65.50)
            (55,65.80)
            (60,65.53)
            (65,64.56)
            (70,66.44)
            (75,66.19)
            (80,65.64)
            (85,66.98)
            (90,65.52)
            (95,65.47)
            (100,65.24)
        };
      \addlegendentry{$tiered$ 5-1}
        
      \addplot[smooth,black, line width = 1.2pt] plot coordinates {
            (0,58.29)
            (5,73.73)
            (10,77.30)
            (15,79.20)
            (20,80.47)
            (25,80.60)
            (30,80.30)
            (35,80.05)
            (40,81.91)
            (45,81.17)
            (50,82.03)
            (55,81.77)
            (60,82.04)
            (65,82.40)
            (70,81.48)
            (75,81.10)
            (80,81.73)
            (85,81.27)
            (90,82.25)
            (95,81.60)
            (100,81.69)
        };
      \addlegendentry{$tiered$ 5-5}
        
      \end{axis}
\end{tikzpicture} 
\label{fig:sub2}}
\caption{Accuracy of the validation set during training.
   (a) Accuracy of the second stage of Meta-Baseline on the validation set of $mini$ImageNet and $tiered$ImageNet along with the training process;
   and (b) accuracy  of our  \name~ method on the validation set of $mini$ImageNet and $tiered$ImageNet along with the training process. 5-1 denotes the 5-way 1-shot problem and 5-5 denotes the 5-way 5-shot problem.}
\label{fig:compare}
\end{figure*}
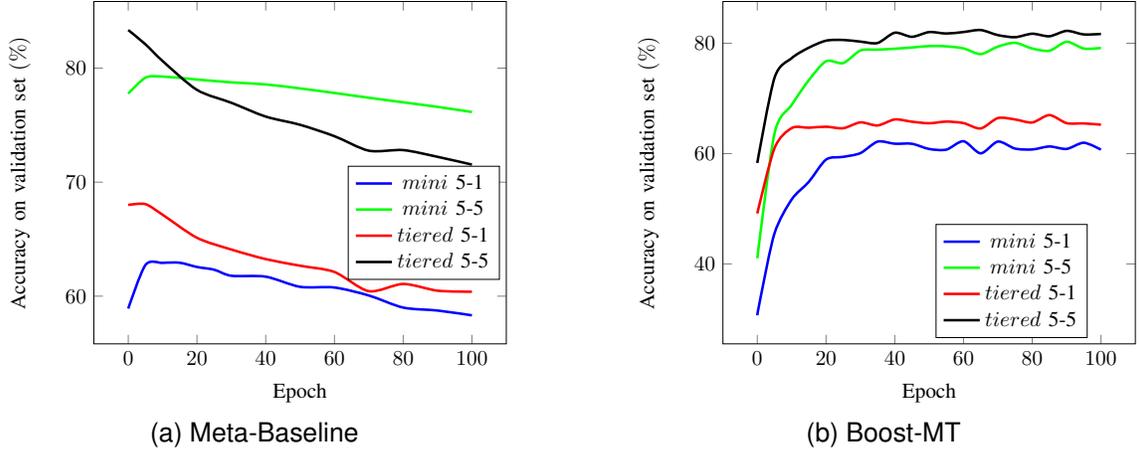

\subsection{Comparison with the baslines}
Following the standard setting of few-shot learning, we conduct experiments on $mini$ImageNet and $tiered$ImageNet, and the results are shown in Tables \ref{tab:miniImage} and \ref{tab:tiered}, respectively. On the two datasets, 
we can observe that the proposed method of Boost-MT consistently achieves competitive results compared to the state-of-the-art Meta-Baseline method on both the 5-way 1-shot and 5-way 5-shot tasks.
Among them, for the $mini$ImageNet dataset, \name~ outperforms Meta-Baseline by $0.94\%$ and $1.74\%$ on 5-way 1-shot and 5-way 5-shot, respectively; for the $tiered$ImageNet dataset, \name~ outperforms Meta-Baseline by $1.11\%$ and $1.17\%$ on 5-way 1-shot and 5-way 5-shot, respectively. 

As shown in Table ~\ref{tab:tiered}, the second stage of Meta-Baseline can hardly gain improvement from the pre-training stage on $tiered$ImageNet, indicating that Meta-Baseline does not play the role of meta-learning on a larger dataset. 
Our method outperforms ProtoNets, Classifier-Baseline and Meta-Baseline, 
showing that 
we make good use of both base class information and meta-learning. 
It is worth mentioning that, in order to make a fair comparison with Meta-Baseline, we only change the training methods according to their settings and also do not use other more sophisticated structural designs or modules.

In addition, when the outer loss is directly used to update the model parameters and remove the inner loop, our method will degenerate to the pre-training method~\cite{baseline++}. When removing the outer loop, our method is a meta-learning method similar to Prototypical Networks~\cite{PrototypicalNetwork}. From the experimental results, we can see that our results are superior to the two baseline methods.

\subsection{Detailed comparison with Meta-Baseline}

This subsection provides a detailed comparison with the current strongest method of Meta-Baseline. 
We first compare the performance of the Meta-Baseline and our approach to the validation set and then analyze the reasons for the superior performance of our model. We then describe the inconvenience of the two-stage training approach in the Meta-Baseline. Finally, we compared the performance of the Meta-Baseline with that of our approach in a cross-domain scenario.

First, we reproduce the Meta-Baseline and draw the performance on the validation classes of $mini$ImageNet and $tiered$ImageNet during the meta-learning stage. 
As shown in Figure \ref{fig:sub1}, although the meta-learning stage improves the model performance on the validation set during the first few epochs, the model performance continues to decline after that. This phenomenon indicates that the two-stage pattern is prone to overfitting, which limits meta-learning to further improve the model performance.

We also draw the performance on validation classes of our model during training.
As shown in Figure \ref{fig:sub2}, our model achieves good performance and converges quickly.
Our model begins to converge at the 40$th$ epoch on $mini$ImageNet and keeps relatively stable fluctuations after that. Also, on the $tiered$ImageNet dataset, the model begins to converge at the 35$th$ epoch. Compared with Meta-Baseline and many other few-shot learning methods~\cite{MAML,ANIL,BOIL}, the convergence is faster and more stable, indicating that our model is more consistent with the paradigm of deep learning.  

\begin{figure}
\centering
\subfloat[On $mini$Imagenet]{
\begin{tikzpicture}[scale=0.48] 

          \begin{axis}[
            xlabel=Epoch, 
            ylabel=Accuracy $(\%)$, 
            tick align=inside, 
            legend style={at={(0.75,0.22))},anchor=north} 
            ]
        
        \addplot[dashed,smooth,blue, line width = 1.5pt] plot coordinates { 
            (0,80.48)
            (5,81.38)
            (10,81.50)
            (15,81.58)
            (20,81.31)
            (25,81.30)
            (30,80.79)
            (40,80.73)
            (50,80.07)
            (60,79.35)
            (70,78.89)
            (80,77.77)
            (90,77.55)
            (100,77.12)
        
        };
          \addlegendentry{Meta-Baseline}
        
          \addplot[dashed,smooth,brown, line width = 1.5pt] plot coordinates { 
            (0,9.44)
            (5,27.04)
            (10,40.84)
            (15,50.34)
            (20,59.43)
            (25,64.81)
            (30,74.10)
            (40,79.03)
            (50,81.35)
            (60,83.65)
            (70,84.50)
            (80,84.58)
            (90,85.78)
            (100,85.82)
        
        };
          \addlegendentry{Boost-MT}
        
          \end{axis}
\end{tikzpicture}
\label{fig:sub11}}
\hfil
\subfloat[On $tiered$Imagenet]{
\begin{tikzpicture}[scale=0.48] 

      \begin{axis}[
            xlabel=Epoch, 
            ylabel=Accuracy $(\%)$, 
            tick align=inside, 
            legend style={at={(0.75,0.22))},anchor=north} 
            ]
        
        \addplot[dashed,smooth,blue, line width = 1.5pt] plot coordinates { 
            (0,71.78)
            (5,65.40)
            (10,63.03)
            (20,62.97)
            (30,62.80)
            (40,62.55)
            (50,62.13)
            (60,62.57)
            (70,62.53)
            (80,62.58)
            (90,62.91)
            (100,62.81)
        };
          \addlegendentry{Meta-Baseline}
        \addplot[dashed,smooth,brown, line width = 1.5pt] plot coordinates { 
            (0,2.57)
            (5,10.47)
            (10,20.67)
            (20,44.16)
            (30,69.09)
            (40,76.96)
            (50,80.73)
            (60,83.23)
            (70,85.66)
            (80,85.77)
            (90,86.00)
            (100,87.66)
        
        };
          \addlegendentry{Boost-MT}
        
      \end{axis}
\end{tikzpicture} 
\label{fig:sub22}}
\caption{The conventional classification accuracy of Meta-baseline (second stage) and our Boost-MT along with the training process.
   }
\label{fig:both-conventional}
\end{figure}
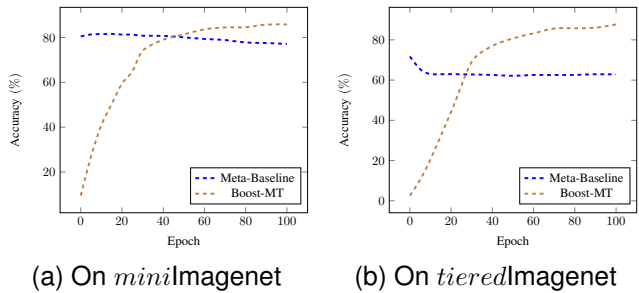

Second, we substantiate the effectiveness of our approach through evaluation on conventional classification tasks throughout the training phase. The results are depicted in Figure \ref{fig:both-conventional}, where results of the second stage of the Meta-Baseline are the same as those in Figure \ref{fig:conventional}. 
In contrast, our model demonstrates a progressive improvement in its performance on conventional classification tasks. Once it surpasses the Meta-Baseline, our model exhibits a trend towards convergence in training accuracy for few-shot tasks as well. 
Different from the Meta-Baseline, the class transferability of our model continues to improve while adapting to few-shot tasks, which is also the reason why our method performs better.
\begin{table}[t]
\caption{Comparison of meta-learning performance under different checkpoints from pre-training of Meta-Baseline on $mini$ImageNet under 5-way 5-shot setting.}
\label{tab:dif-classifier}
\begin{center}
    \begin{tabular}{|l| c| c| c| c| c|}
    \hline
    \bf Epoch  & \bf 91 & \bf 93  & \bf 95 & \bf 96 & \bf 97\\      
    \hline
    Pre-training &77.30 & 78.36  & 78.91 &79.25 & 78.67\\
    \hline
    Meta-training  & 78.74 & 79.25 & 78.55 & 79.41 & 79.49 \\
    \hline
    \end{tabular}
\end{center}
\end{table}

Third, to describe the inconveniences of Meta-Baseline, we select 5 checkpoints when the pre-training is close to 100 epochs, and continue the second stage of meta-training after each checkpoint. As shown in Table \ref{tab:dif-classifier}, we find that the training results performing best on validation in the first stage are not the best in the second stage. Therefore, if we want the best final result of Meta-Baseline, it is better to try multiple first-stage checkpoints, which illustrates the inconveniences of two-stage training.

\begin{table}[t]
\caption{Comparison with Meta-Baseline for 5-way classification in cross-domain scenarios. 
$\dagger$: reproduced using our setting.}
\label{tab:cross-domain}
\begin{center}
    \begin{tabular}{|l|c|c|c|}
    \hline
    \bf Model & \bf Backnone & \bf 1-shot & \bf 5-shot\\
    \hline
    Meta-Baseline$^\dagger$ & ResNet-12 & 44.69 ± 0.59 &  61.60 ± 0.54\\
    \hline
    Boost-MT (ours)  & ResNet-12 & 45.18 ± 0.60 & 62.91 ± 0.57\\
    \hline
    \end{tabular}
\end{center}

\end{table}

Forth, we further evaluate our model in cross-domain scenarios. We train our model on $mini$ImageNet and then test on CUB. The results in Table \ref{tab:cross-domain} show that our model outperforms Meta-Baseline by $1\%$, 
indicating that our model has stronger domain transfer capability. 

\subsection{Ablation study}
In this subsection, we perform ablation experiments to analyze how each component affects the performance of our method and 
do comparison 
under 5-way setting on $mini$ImageNet using the ResNet-12 backbone. We mainly study the following two aspects: the effectiveness of our two loops setting and the rationality of our update method on the feature extractor and classification header. 

\begin{table}[t]
\caption{Comparison to methods without inner or outer loop guidance on $mini$ImageNet. 
We show the average 5-way accuracy with 95$\%$ confidence interval.}
\label{tab:without-svrg}
\begin{center}
    \begin{tabular}{|l|c|c|c|}
    \hline
    \bf Model & \bf Backnone & \bf 1-shot & \bf 5-shot\\
    \hline
    Meta-Baseline & ResNet-12 & 63.17 ± 0.23 & 79.26 ± 0.17\\
    \hline
    Boost-MT (w/o inner) & ResNet-12 & 62.06 ± 0.62 & 80.12 ± 0.44\\
    Boost-MT (w/o outer) & ResNet-12 & 60.88 ± 0.67 & 78.19 ± 0.50\\
    \hline
    Boost-MT (all) & ResNet-12 & 64.01 ± 0.97 & 81.00 ± 0.59\\
    \hline
    \end{tabular}
\end{center}
\end{table}

First, we investigate the effect of the inner loop and the outer loop of our method.  
We first remove the inner loop of our method and only the loss of the outer loop is used to guide the parameters update of the feature extractor and linear layer. Second, we remove the outer loop, and the parameters of the feature extractor are directly updated by calculating the loss of the query set in the inner loop. As shown in Table \ref{tab:without-svrg}, when either the inner loop or the outer loop is removed, the model shows a significant decline in performance, which is even below that of the Meta-Baseline, and only when the two cooperate with each other, the performance of the model is significantly higher.
This further shows that using gradient information from the base class is more beneficial to the meta-training of the model than the parameter weight.
\begin{table}[t]
\caption{Comparison to two methods, variant 1 and variant 2, on $mini$ImageNet. 
We show the average 5-way accuracy with 95$\%$ confidence interval.}
\label{tab:variant}
\begin{center}
    \begin{tabular}{|l|c|c|c|}
    \hline
    \bf Model & \bf Backnone & \bf 1-shot & \bf 5-shot\\
    \hline
    Boost-MT-f & ResNet-12 & 63.29 ± 0.88  & 80.44 ± 0.60\\
    Boost-MT-c
    & ResNet-12 & 59.62 ± 0.90 & 78.56 ± 0.61\\
    \hline
    Boost-MT  & ResNet-12 & 64.01 ± 0.97 & 81.00 ± 0.59\\
    \hline
    \end{tabular}
\end{center}
\end{table}

Second, we investigate the rationality of our update method on the feature extractor and classification header. 
In the course of training, the outer loop exclusively updates the classification header, while the inner loop is dedicated solely to updating the feature extractor. To substantiate the rationale behind, we devise two variant methods. In Boost-MT-f, we update the feature extractor both in inner and outer loops. In Boost-MT-c, we update the classification header both in both loops. As depicted in Table \ref{tab:variant}, both methods exhibit inferior performance compared to ours. In our framework, the outer loop primarily addresses the conventional classification task, while the inner loop is designed for handling few-shot classification tasks. By transferring the loss from the outer loop to the inner loop and updating parameters concurrently with the inner loop loss, the feature extractor can preserve its proficiency in traditional image classification while adapting to few-shot tasks. Consequently, the feature extractor only requires parameter updates in inner loop.
Similarly, the classification header is utilized to assess performance of the feature extractor on the traditional classification task after the inner loop update and generate new outer loss. As a result, there is no necessity to further update the classification header in inner loop.
\begin{table}[t]
\caption{The average 5-way 5-shot performance of our model under different number of inner loops per outer loop on $mini$ImageNet.}
\label{tab:num_of_inner_loop}
\begin{center}
    \begin{tabular}{|c|c|c|c|c|c|}
    \hline
    \bf $T$ & \bf 1 & \bf 5 & \bf 10 & \bf 15 & \bf 20\\
    \hline
    Accuracy & 76.41 & 77.64 & 79.66 & 78.57 & 77.72\\
    \hline
    \end{tabular} 
\end{center}
\end{table}

\subsection{Parameter study}
We further study the impact of different numbers of inner loops per outer loop. The results are shown in Table \ref{tab:num_of_inner_loop}. When the number is either too small or too big, the performance of the model will decline. The performance is good when the number of inner loops is between 5 and 15. In our final experiment, the number of inner loops is set to 10. 

In the outer loop, the feature extractor and classifier are utilized to calculate the classification loss to achieve class transferability.
In the inner loop, the feature extractor and metric module work together to calculate the loss for each specific few-shot classification task. This process allows the model to adapt and fine-tune its parameters specifically for the given few-shot classification task.

We speculate that the number of inner loops mainly affects the update of internal parameters. When the number of internal tasks is too small, the number of parameter iterations is less, and the meta-training is not sufficient. When the number of inner loops is too large, the information of the base class could not be used effectively.


\subsection{Possible extension of our method}
The proposed method introduces a novel end-to-end training framework, characterized by a model-agnostic training process. As a result, the proposed method should exhibit broad applicability to existing meta-learners and potential future learning algorithms. To validate the versatility of our approach, we conduct additional experiments on two existing methods. 
CAN \cite{can} introduces a Cross Attention Module to generate cross attention maps for each pair of class features and query sample features, thereby emphasizing the target object regions and enhancing the discriminative nature of the extracted features. The model is updated using a combination of the global classification loss and few-shot task loss as the overall loss function. In our scenario, we employ the global classification loss as the outer loss and the few-shot task loss as the inner loss.
FEAT \cite{FEAT} proposes an innovative approach to adapt instance embeddings to the target classification task using a set-to-set function, resulting in task-specific and discriminative embeddings. During training, it utilizes standard meta-learning with pre-training. In our case, we directly apply our framework.

Table \ref{tab:extension} illustrates the gains achieved by integrating Boost-MT into each method. Notably, we observe an average increase of approximately $1\%$ after incorporating Boost-MT. This underscores the adaptability of our method across multiple metric-based systems. Furthermore, these gains consistently persist across all methods, regardless of the initial performance level. 
The findings provide further evidence for the hypothesis that our method can be seamlessly integrated with established meta-learning approaches.


\begin{table}[!t]
\caption{Comparison to methods with our Boost-MT training framework on $mini$ImageNet. 
We show the average 5-way accuracy with 95$\%$ confidence interval.}
\label{tab:extension}
\begin{center}
    \begin{tabular}{|l|c|c|c|}
    \hline
    \bf Model & \bf Backnone & \bf 1-shot & \bf 5-shot\\
    \hline
    CAN~\cite{can} & ResNet-12 & 63.85 ± 0.48 & 79.44 ± 0.34\\
    \hline
    CAN+Boost-MT & ResNet-12 & 64.83 ± 0.52 & 80.84 ± 0.59\\
    \hline
    FEAT~\cite{FEAT} & ResNet-12 & 66.78 & 82.05\\
    \hline
    FEAT+Boost-MT & ResNet-12 & 68.73 ± 0.61 & 83.00 ± 0.58\\
    \hline
    \end{tabular}
\end{center}
\end{table}

\section{Conclusion}
In this work, we propose a new few-shot learning framework termed Boost-MT, which boosts meta-training with base class information.
Our experiments indicate that simply concatenating pre-training and meta-training like Meta-Baseline can not make full use of the base class information, because it is difficult for meta-training to measure the fixed representation from pre-training. 
As an end-to-end framework, our method leverages the gradient information from the base class to guide meta-learning, avoid mutual subtraction of pre-training and meta-training, converge quickly, and outperform the Meta-Baseline.
To investigate the efficacy of the inner loop and outer loop, we conduct two additional experiments in which either the inner loop or the outer loop is excluded from the model update process. In these experiments, the model is trained using only one of the loops.
The performance of the derived methods is subpar, indicating that the omission of pre-training holds informative value for the meta-training process. However, it is unwise to solely aggregate the two stages or treat them as distinct entities.
Comprehensive experimental results show that our proposed method achieves competitive performance on two well-known datasets, $mini$ImageNet and $tiered$ImageNet.

We will investigate possible improvement on Boost-MT, as a new framework for meta-training, in our future work. Also, our experiments have demonstrated that the base class information is useful for meta-learning, that also deserves further exploration. Moreover, because our work involves a training process that is algorithm-agnostic, the potential benefits of applying the training process to a wide range of meta-learners are worth exploring in the future.

\section*{Acknowledgments}
This work is supported by National Natural Science Foundation (U22B2017).




\bibliographystyle{IEEEtran}
\bibliography{References}

\newpage

 

\begin{IEEEbiography}[{\includegraphics[width=1in,height=1.25in,clip,keepaspectratio]{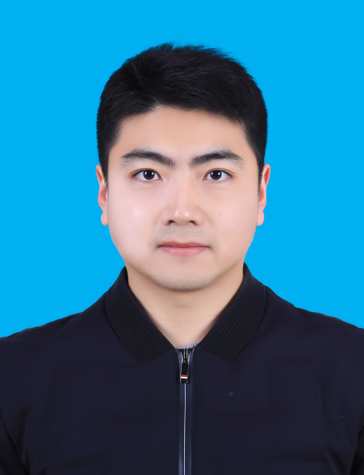}}]{Weihao Jiang}
is currently pursuing a Ph.D. in School of Computer Science and Technology, Huazhong University of Science and Technology, Wuhan, China.
His research focuses on deep learning,
few-shot learning and meta-learning.
\end{IEEEbiography}

\vspace{11pt}

\begin{IEEEbiography}[{\includegraphics[width=1in,height=1.25in,clip,keepaspectratio]{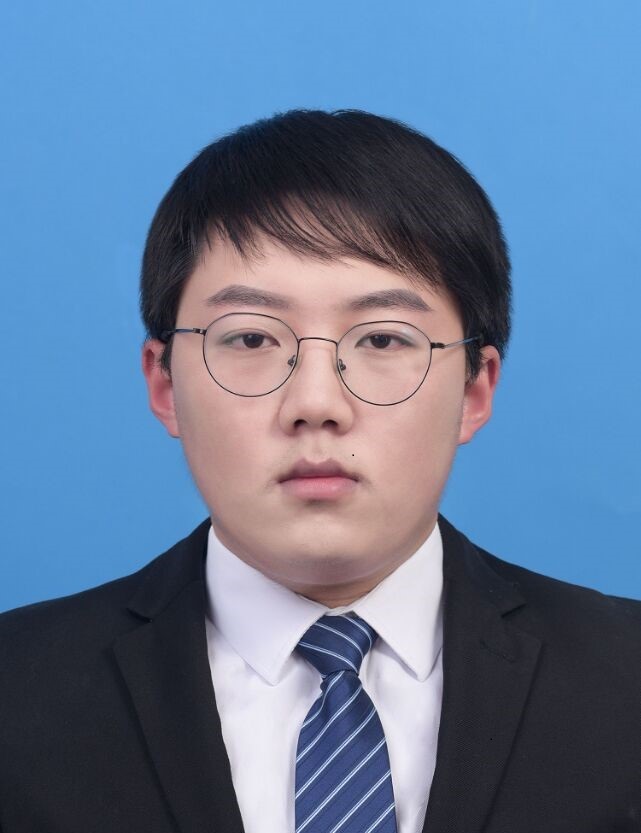}}]{Guodong Liu}
received the master degree in computer application technology from the Huazhong University of Science and Technology, Wuhan, China, in 2023. 
His research mainly focuses on deep learning, few-shot learning, and meta-learning.
\end{IEEEbiography}

\vspace{11pt}
\begin{IEEEbiography}[{\includegraphics[width=1in,height=1.25in,clip]{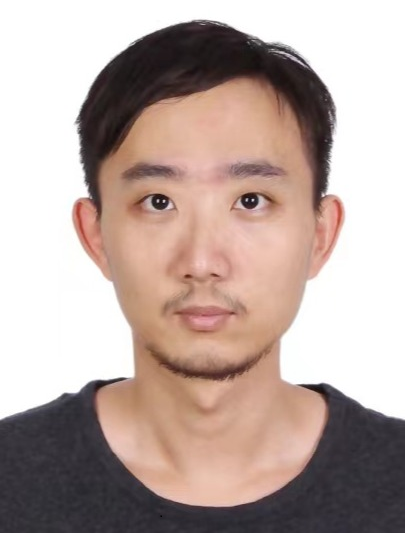}}]{Di He} 
is currently an assistant professor at Peking University. He was previously a Senior Researcher in the Machine Learning Group at Microsoft Research Asia. He obtained his bachelor, master and Ph.D. degrees from Peking University.
Di’s main research (current and past) includes representation learning (large-scale pre-training of foundation models, graph representation learning) and trustworthy machine learning. 
\end{IEEEbiography}

\begin{IEEEbiography}[{\includegraphics[width=1in,height=1.25in,clip,keepaspectratio]{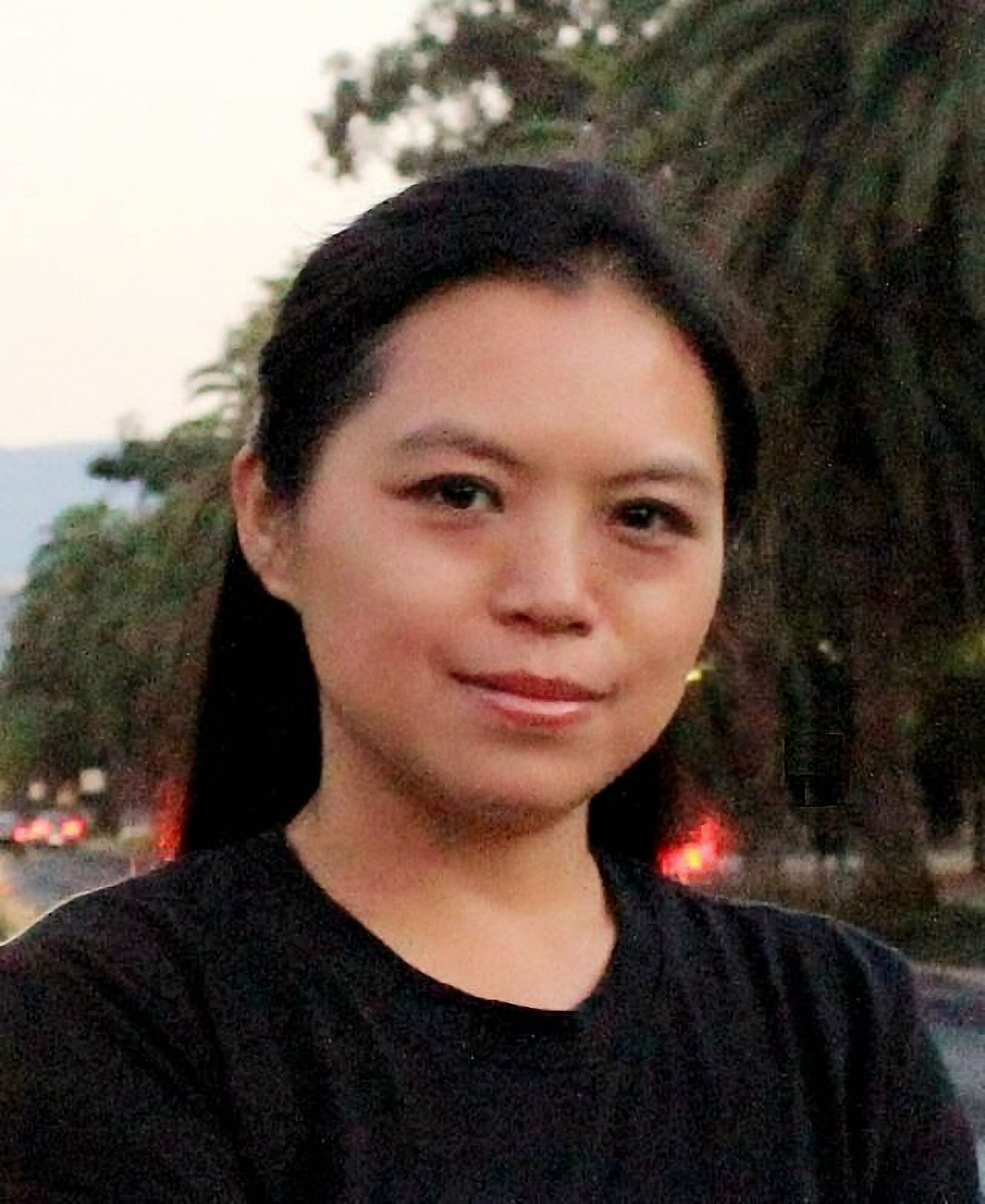}}]{Kun He}
(SM18) is currently a Professor in School of Computer Science and Technology, Huazhong University of Science and Technology, Wuhan, P.R. China. She received the Ph.D. degree in system engineering from Huazhong University of Science and Technology, Wuhan, China, in 2006. She had been with the Department of Management Science and Engineering at Stanford University in 2011-2012 as a visiting researcher. She had been with the department of Computer Science at Cornell University in 2013-2015 as a visiting associate professor, in 
2016 as a visiting professor, and in 2018 as a visiting professor. She was honored as a Mary Shepard B. Upson visiting professor for the 2016-2017 Academic year in Engineering, Cornell University, New York. Her research interests include adversarial machine learning, representation learning, graph data mining, and combinatorial optimization. 
\end{IEEEbiography}



\vfill

\end{document}